\documentclass[11pt, letterpaper]{article}

\usepackage[left=1in, right=1in, top=1in, bottom=1in]{geometry}
\usepackage{bm}
\usepackage{type1cm}
\usepackage{lettrine}
\usepackage{amsmath,amssymb,amsthm}
\usepackage{moreverb}
\usepackage{mathtools}
\usepackage{algorithmic}
\usepackage{graphics}
\usepackage{graphicx}
\usepackage{caption}
\usepackage{extarrows}
\usepackage{color}
\usepackage{framed}
\usepackage{wrapfig}
\usepackage{mathrsfs}
\usepackage{multirow}
\usepackage{longtable}
\usepackage{hyperref}
\usepackage{paralist}
\usepackage{indentfirst}
\usepackage{relsize}
\usepackage{upgreek}
\usepackage[dvipsnames,table]{xcolor}
\usepackage{booktabs}
\usepackage{authblk}
\usepackage{subcaption}
\usepackage{threeparttable}
\usepackage[sort&compress,numbers]{natbib}
\usepackage[figurename=Figure]{caption}
\usepackage[normalem]{ulem}
\usepackage{tcolorbox}

\graphicspath{ {./Figure/} }
\usepackage[font=footnotesize,labelfont=bf]{caption}

\providecommand{\keywords}[1]{\textbf{\textit{Keywords: }} #1}

\hypersetup{
bookmarks=true,
bookmarksopen=true,
bookmarksnumbered=true,
unicode=true,
pdftoolbar=true,
pdfmenubar=true,
pdffitwindow=false,
pdfstartview={FitH},
pdftitle={S-Researcher},
pdfauthor={},
pdfsubject={},
pdfcreator={},
pdfproducer={},
pdfkeywords={LLM, social science, simulation, research paradigm},
pdfnewwindow=true,
colorlinks=true,
linkcolor=blue,
citecolor=blue,
filecolor=blue,
urlcolor=blue
}

\begin{document}
% \linenumbers
\title{\textbf{LLM Agents as Social Scientists: A Human–AI Collaborative Platform for Social Science Automation}}

\author[1,2]{Lei Wang}
\author[1,2]{Yuanzi Li}
\author[1,2]{Jinchao Wu}
\author[1,2]{Heyang Gao}
\author[1,2]{Xiaohe Bo}
\author[1,2]{Xu Chen}
\author[1,2]{Ji-Rong Wen}

\affil[1]{\small Gaoling School of Artificial Intelligence, Renmin University of China, Beijing, China}
\affil[2]{Beijing Key Laboratory of Big Data Management and Analysis Methods, Beijing, China}
% \affil[*]{Corresponding authors}

\date{}

\maketitle

\normalsize

\vspace{-18pt}
\begin{abstract}
    Traditional social science research often requires designing complex experiments across vast methodological spaces and depends on real human participants, making it labor-intensive, costly, and difficult to scale.
    Here we present S-Researcher, an LLM-agent-based platform that assists researchers in conducting social science research more efficiently and at greater scale by ``siliconizing'' both the research process and the participant pool.
    To build S-Researcher, we first develop YuLan-OneSim, a large-scale social simulation system designed around three core requirements: \textit{generality} via auto-programming from natural language to executable scenarios, \textit{scalability} via a distributed architecture supporting up to 100,000 concurrent agents, and \textit{reliability} via feedback-driven LLM fine-tuning.
    Leveraging this system, S-Researcher supports researchers in designing social experiments, simulating human behavior with LLM agents, analyzing results, and generating reports, forming a complete human--AI collaborative research loop in which researchers retain oversight and intervention at every stage.
    We operationalize LLM simulation research paradigms into three canonical reasoning modes (induction, deduction, and abduction) and validate S-Researcher through systematic case studies: inductive reproduction of cultural dynamics consistent with Axelrod's theory, deductive testing of competing hypotheses on teacher attention validated against survey data, and abductive identification of a cooperation mechanism in public goods games confirmed by human experiments.
    S-Researcher establishes a new human--AI collaborative paradigm for social science, in which computational simulation augments human researchers to accelerate discovery across the full spectrum of social inquiry.
\end{abstract}

\keywords{large language model, social simulation, computational social science}

\vspace{12pt}
\section*{Introduction}

Social science has long advanced human civilization by providing deep insights into human behavior, social structures, and cultural dynamics.
Traditional research typically begins with the design of methodological frameworks, such as surveys~\cite{gideon2012handbook,wright2010survey}, controlled experiments~\cite{falk2009lab,baldassarri2017field}, and longitudinal studies, followed by data collection from human participants to derive findings~\cite{caruana2015longitudinal}.
However, this process confronts two structural bottlenecks that fundamentally constrain the pace, scale, and depth of social science research.
From the researcher's perspective, the methodological design space is vast, rendering the identification of optimal experimental configurations and exhaustive evaluation of competing hypotheses both time-consuming and, in practice, often intractable; moreover, hypothesis generation and appraisal remain heavily dependent on individual intuition — a process that is inherently slow, idiosyncratic, and difficult to scale. From the participant's perspective, the prohibitive costs and ethical constraints of human recruitment fundamentally limit experimental scale, precluding the capacity to capture population-level dynamics, isolate causal mechanisms, or systematically explore broad parameter spaces~\cite{squazzoni2014social,heath2009survey,lazer2009computational}. Collectively, these bottlenecks have entrenched a persistent divide between the scope of social theory and the practical capacity to subject it to rigorous empirical test.

The emergence of large language models (LLMs) has opened a new frontier for addressing these challenges.
Recent work has demonstrated that LLM-based agents can produce human-like behaviors across diverse social contexts~\cite{wang2024survey,argyle2023out,aher2023using,gao2024large}, including daily life~\cite{park2023generative}, education~\cite{zhang2024simulating}, social networks~\cite{zhang2024generative,wang2025user,zhang2024trendsim}, opinion dynamics~\cite{chuang2023simulating,hu2025simulating}, and elections~\cite{zhang2024electionsim}.
Although promising, existing approaches focus primarily on using LLMs as ``silicon participants,'' replacing human subjects in isolated, scenario-specific settings.
A systematic platform that augments human researchers across the entire workflow, spanning hypothesis generation, experiment design, simulation, and analysis, while preserving human oversight and domain expertise, remains largely unexplored.

\begin{figure}[t!]
    \centering
    \setlength{\fboxrule}{0.pt}
    \setlength{\fboxsep}{0.pt}
    \includegraphics[width=1\linewidth]{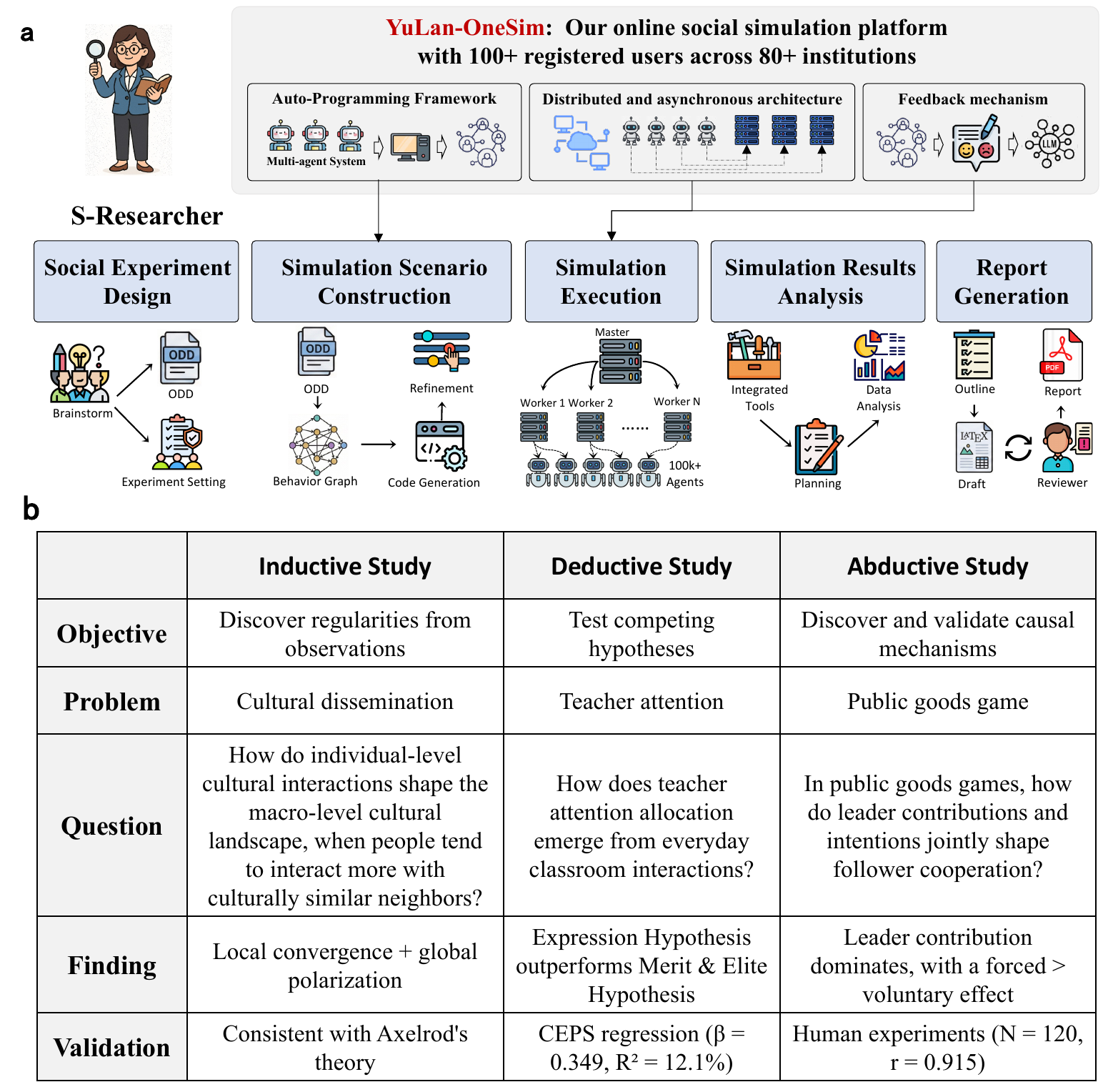}
\caption{Overview of S-Researcher.
(a) The complete workflow of S-Researcher: users input their research topics, after which simulation scenarios are automatically constructed, executed, and summarized into comprehensive reports. Researchers can intervene at every stage.
(b) Summary of three case studies organized by reasoning paradigm: induction, deduction, and abduction.}
    \label{fig:overall}
\end{figure}

To bridge this gap, we introduce S-Researcher, an agentic platform that augments human researchers by siliconizing both the research process and the participant pool. It supports the full research loop, covering experimental design, human behavior simulation with LLM agents, result analysis, and report generation, while enabling researchers to intervene and steer the process at every stage (Fig.~\ref{fig:overall}).
While recent years have seen the emergence of numerous AI-driven research systems in the natural sciences~\cite{lu2024ai,ren2025towards,wang2023scientific}, these systems are not directly transferable to the social sciences, which typically center on human subjectivity rather than objective natural patterns. 
While Manning et al.~\cite{manning2024automated} make an early step toward AI social researchers, several fundamental challenges remain unresolved, most notably, developing a simulation system with sufficient \textit{generality} to accommodate the flexible experimental designs generated by upstream LLMs, achieving the \textit{scalability} necessary for population-level experiments, and ensuring the \textit{reliability} of simulated outcomes to support credible scientific inference.
% Automating social science research requires, as a critical foundation, an effective social simulator capable of faithfully reproducing human behaviors. We identify three essential requirements for such a simulator: \textit{generality}, to accommodate the diverse experimental designs generated by upstream LLMs; \textit{scalability}, to support population-level experiments; and \textit{reliability} of simulated outcomes, to enable credible scientific inference.

To address these challenges, we first develop YuLan-OneSim, the social simulation engine underlying S-Researcher, with three integrated capabilities. An auto-programming framework translates natural-language scenario descriptions into executable simulation code via ODD protocols~\cite{grimm2010odd} and validated behavior graphs, providing the \textit{generality} to support arbitrary experimental designs and freeing researchers from engineering overhead so they can focus on scientific questions. A distributed, event-driven architecture executes agent behaviors in parallel with topology-aware allocation, achieving \textit{scalability} to 100,000 concurrent agents. A Verifier–Reasoner–Refiner–Tuner (VR$^2$T for short) feedback mechanism iteratively evaluates simulation outputs and fine-tunes the backbone LLMs, ensuring the \textit{reliability} required for rigorous social-scientific inquiry. Extensive experiments across 50 simulation scenarios spanning eight social-science domains demonstrate the generality, scalability, and reliability of our simulator. To broaden accessibility, we have deployed YuLan-OneSim as an online service\footnote{\url{http://www.yulan-onesim.cn/}}, which has attracted over {100+ registered users across 80+ institutions}, enabling researchers to design and run social simulations directly through a web interface without local infrastructure.

Leveraging YuLan-OneSim, S-Researcher converts user-defined research question and scenario descriptions into experimental designs, executes simulations, collects and analyzes quantitative data, and generates comprehensive reports. We organize this workflow around the three canonical modes of scientific reasoning~\cite{peirce1931collected}: induction, deduction, and abduction. For inductive research, the system executes large-scale simulations and identifies emergent patterns from agent interactions. For deductive research, competing hypotheses are encoded as distinct configurations and tested simultaneously. For abductive research, the system generates candidate mechanisms and designs validation experiments. S-Researcher operationalizes all three paradigms into an integrated, human-in-the-loop workflow: researchers can modify generated designs, inject domain expertise, supply custom data, or use individual modules independently, ensuring that the platform serves as a collaborative accelerator rather than a black-box replacement.
We validate the framework through three case studies.
In the inductive paradigm, the S-Researcher autonomously designs and executes a cultural dissemination experiment that successfully reproduces the convergence and polarization dynamics predicted by Axelrod's classical theory~\cite{axelrod1997dissemination}, with LLM-driven agents whose behaviors emerge from language model reasoning rather than hard-coded rules.
In the deductive paradigm, large-scale classroom simulation (221 classrooms, 5,525 student agents) simultaneously tests three competing hypotheses about teacher attention allocation, with results directionally validated against the China Education Panel Survey(CEPS)~\cite{national2014data}.
In the abductive paradigm, the most demanding test, S-Researcher identifies a leader demonstration mechanism in public goods games, which is independently confirmed by parallel human experiments ($N = 120$) with a cross-condition correlation of $r = 0.915$.
These results demonstrate that LLM simulation, when embedded within a human--AI collaborative workflow, can meaningfully accelerate social science research.
By providing a general, scalable, and reliable platform that substantially lowers the technical barrier for computational social science while keeping researchers in the loop, S-Researcher opens a new avenue for augmenting human expertise in the social sciences.

\section*{Results}

\subsection*{YuLan-OneSim: a general, scalable, and reliable platform for social simulation}

S-Researcher requires a simulation platform that can execute diverse experimental designs, scale to population-level experiments, and produce trustworthy behavioral outputs. We developed YuLan-OneSim to meet these requirements through three integrated capabilities: an auto-programming framework for generality, a distributed simulation architecture for scalability, and a feedback mechanism for reliability.

\begin{figure}[t!]
    \centering
    \setlength{\fboxrule}{0.pt}
    \setlength{\fboxsep}{0.pt}
    \includegraphics[width=\linewidth]{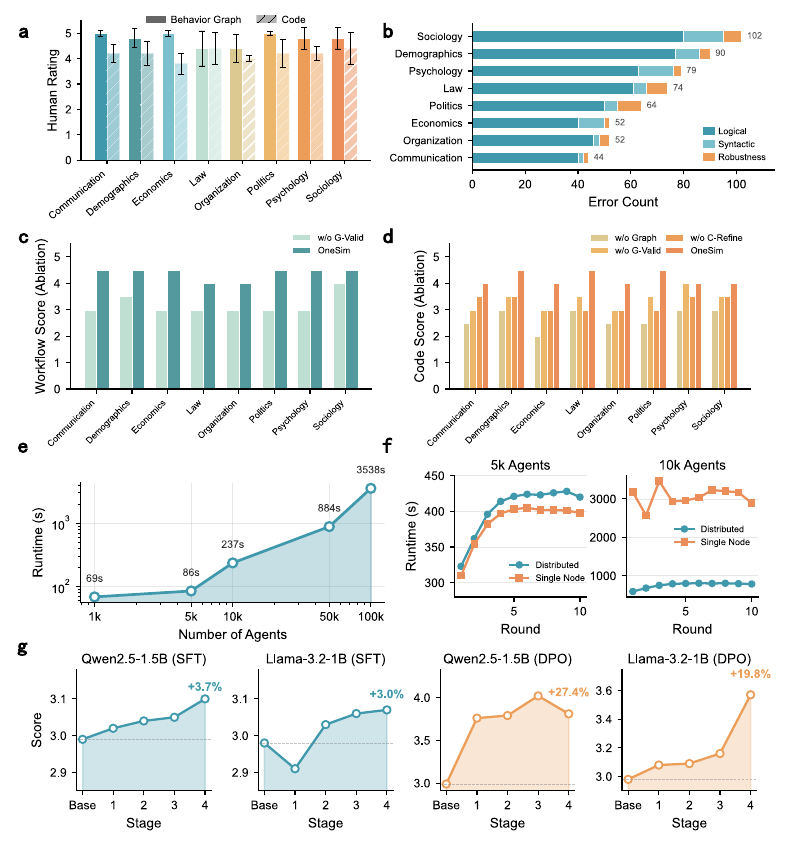}
    \caption{\textbf{Platform capability validation of YuLan-OneSim.}
\textbf{a}, Human expert ratings for auto-generated code across eight social science domains.
\textbf{b}, Error type distribution across domains.
\textbf{c}, Ablation study on workflow scores: OneSim vs.\ w/o G-Valid.
\textbf{d}, Code quality ablation: four variant comparison.
\textbf{e}, Runtime scaling with agent count.
\textbf{f}, Distributed vs.\ single-node deployment efficiency.
\textbf{g}, Feedback-driven optimization trajectories for two backbone models (Qwen2.5-1.5B, Llama-3.2-1B) under SFT and DPO strategies.}
    \label{fig:onesim}
\end{figure}

\textbf{Auto-programming framework for generalizable simulation.}
To support arbitrary experimental designs generated by S-Researcher's upstream modules, YuLan-OneSim translates natural language scenario descriptions into executable simulation code through a structured pipeline.
User inputs are first formalized following the ODD (Overview, Design Concepts, Details) protocol~\cite{grimm2010odd}, with an interactive agent resolving ambiguities.
A behavior graph is then constructed to represent agent actions and interactions, where the system can automatically validate its structural consistency.
Guided by this graph, the final simulation code is generated and iteratively refined to ensure correctness.
To demonstrate the capability of our auto-programming framework, we evaluate it on agent behavior graph and simulation code generation across 50 scenarios spanning eight social-science domains, including communication, demography, economics, law, organizational studies, political science, psychology, and sociology.
We recruit {3} domain experts to rate the generated outputs on a 5-point Likert scale~\cite{likert1932technique} with explicit criteria for each level. As shown in Fig.~\ref{fig:onesim}a, agent behavior graphs consistently approach perfect scores (5.0/5.0), while the average score of the simulation code is about {4.2} - a strong result given that the automated generation completes in minutes what would otherwise require hours of manual programming.
For a closer examination of failure modes, we categorize the three major error types in the generated code in Fig.~\ref{fig:onesim}b. Logical errors such as missing variable initialization, redundant conditionals, and incorrect string parsing dominate across most domains. Notably, such errors are largely amenable to automated detection and correction by existing static code analysis tools, suggesting a practical pathway for further improving generation quality.
Finally, we conduct an ablation study by individually removing each component of our framework (Fig.~\ref{fig:onesim}c–d). Here, G-Refine and C-Refine denote the LLM-driven validation and iterative repair modules applied at the behavior graph stage and the code generation stage, respectively. For behavior graph generation, removing G-Refine causes a 27.1\% performance drop, confirming that automated verification is essential for producing high-quality behavior graphs. For code generation, the agent behavior graph is by far the most influential component—its removal leads to the largest degradation by 35.8\%, validating our core design of progressively translating natural-language descriptions into structured behavior graphs before generating executable code. Removing C-Refine and G-Refine also results in meaningful drops of 22.4\% and 19.4\%, respectively. Notably, G-Refine's impact on code quality reflects the fact that the behavior graph serves as the direct input to code generation, and lower graph quality inevitably propagates downstream.

% Existing LLM-based simulation platforms such as GenSim~\cite{gensim}, OASIS~\cite{oasis}, and AgentSociety~\cite{agentsociety} rely on manual scenario programming or restrict users to predefined templates—a fundamental limitation when the simulator must serve as the execution engine for an autonomous AI researcher, which by nature generates novel and unpredictable experimental designs. YuLan-OneSim's auto-programming framework removes this bottleneck by enabling fully automated scenario construction from natural language, making it, to our knowledge, the only simulation platform capable of operating in a closed loop with an upstream AI research agent.

\textbf{Distributed architecture for scalable simulation.}
Large-scale social experiments are essential for achieving statistical power, capturing population-level dynamics, and testing theories across diverse conditions.
YuLan-OneSim employs an event-driven, asynchronous architecture: events encapsulate agent actions and environmental changes, while agents subscribe to relevant event types, enabling natural parallelism.
A Master--Worker distributed design orchestrates parallel execution: the master node manages agent registry and global state; worker nodes execute agent logic in parallel; high-performance gRPC~\cite{grpc2018high} communication with optimized batching minimizes overhead; topology-aware allocation co-locates frequently interacting agents to reduce cross-node communication.
To evaluate the scalability of our framework, we examine two questions: (1) whether our framework can support large-scale LLM-agent simulations, and (2) how much our event-driven parallel architecture contributes to efficiency. 
As shown in Fig.~\ref{fig:onesim}e, the simulator scales smoothly from 1k to 100k agents, completing a single simulation round with 100k concurrent agents in approximately 3{,}538 seconds. 
Fig.~\ref{fig:onesim}f further compares our distributed architecture against a single-node baseline. When agent counts are small, the speedup is modest, as LLM inference dominates the overall cost. However, at 10k agents our framework achieves a 3–4 $\times$ speedup, demonstrating its capacity to support population-level social simulations.

\textbf{Feedback mechanism for reliable simulation.}
General-purpose LLMs are not specifically trained for social-science domains and often produce unreliable simulation outcomes that can accumulate over extended runs. To address this, we design a Verifier–Reasoner–Refiner–Tuner (VR$^2$T) multi-agent feedback framework that iteratively evaluates simulation outputs and fine-tunes the backbone LLMs. The verifier assesses response correctness, the reasoner provides explanatory diagnoses, the refiner revises low-scoring outputs, and the tuner conducts supervised or reinforcement-based fine-tuning. Each component can operate in either fully automatic or human-in-the-loop mode, enabling the platform to serve as a ``social-science gym'' in which LLMs progressively acquire social intelligence through continual simulation and feedback. To validate this mechanism, we adopt two widely used LLMs, Qwen2.5-1.5B~\cite{yang2024qwen2} and Llama-3.2-1B~\cite{grattafiori2024llama3herdmodels}, as backbones and fine-tune them using either supervised fine-tuning (SFT)~\cite{ouyang2022training} or direct preference optimization (DPO)~\cite{rafailov2023direct} based on the system's feedback. We conduct four successive simulation rounds, evaluating reliability after each round via LLM-as-a-Judge. As shown in Fig.~\ref{fig:onesim}g, both training strategies yield consistent improvements. Under SFT, the two models achieve steady gains of 3.0–3.7\%. Under DPO, improvements are substantially larger: Qwen2.5-1.5B improves by 27.4\% and Llama-3.2-1B by 19.8\%, with gains concentrated in the first two to three iterations before plateauing.

\subsection*{S-Researcher: automated social science research across three reasoning paradigms}
Having established YuLan-OneSim as a general-purpose, scalable, and reliable simulation engine, we now describe how we build upon it to create S-Researcher—an end-to-end AI social scientist that automates the complete research workflow from experiment design to report writing.

\textbf{Three modes of social-scientific inquiry.} Following Peirce's classical trichotomy~\cite{peirce1931collected}, we design S-Researcher to support three complementary modes of reasoning that together span the full spectrum of social-science methodology. \textit{Induction} begins without presupposing a theory: the system runs large-scale simulations and applies statistical pattern discovery to identify regularities emerging from agent interactions. \textit{Deduction} operates in the opposite direction—starting from an existing theory, competing hypotheses are encoded as distinct simulation configurations and executed in parallel to assess which theoretical expectations hold. \textit{Abduction} reasons backward from observed phenomena to their underlying causal mechanisms~\cite{schurz2008patterns}: given an unexplained observation, the system designs controlled experiments that systematically manipulate candidate causal variables to decompose the phenomenon into its core drivers. By unifying all three modes within a single platform, S-Researcher offers researchers a complete methodological toolkit for pattern discovery, theory testing, and mechanism inference.

\textbf{Pipeline.} 
In the \textit{experiment design} stage, the system first analyzes the user-provided research question and selects the most appropriate reasoning mode. It then generates a natural-language ODD protocol, which YuLan-OneSim's auto-programming framework automatically translates into executable simulation code. Finally, YuLan-OneSim executes the simulation and returns the experimental results.
In the \textit{results analysis} stage, a multi-agent analytical framework processes the typically large and redundant simulation outputs: a planner agent identifies the most relevant variables, an execution agent conducts cross-condition comparisons using LLM-based reasoning alongside external statistical tools (e.g., t-tests) and visualization, and a reviewer agent evaluates the analyses and provides iterative feedback. In the \textit{report generation} stage, a structured template spanning abstract through conclusions is populated with content drawn from the research topic, simulation scenarios, and analytical results; users may optionally supply a reference report to guide style and structure, and a generation–review cycle iteratively refines the output until it meets a predefined quality threshold.

\textbf{Human–AI collaboration.} To support controllable social-science automation, each module accepts direct intervention from human researchers. For instance, users can modify the generated experimental designs or upload custom configuration files to initialize simulations. Moreover, the modules can be used independently—one may, for example, employ only the results analysis and report generation components by supplying their own experimental data.

In the following sections, we demonstrate the effectiveness of S-Researcher through three case studies, each corresponding to one of the reasoning modes above: an inductive study on cultural dissemination dynamics, a deductive test of competing hypotheses on teacher attention allocation, and an abductive discovery of cooperation mechanisms in public goods games (Fig.~\ref{fig:overall}b).

% \begin{table}[h!]
% \centering
% \caption{\textbf{Three research paradigms derived from $O = F(T, C)$.}}
% \label{tab:paradigms}
% \begin{tabular}{lcccc}
% \toprule
% Paradigm & Form & Core action & Unique simulation advantage & Typical application \\
% \midrule
% Inductive & $(O, C) \rightarrow T$ & Discover regularities & Scale $\times$ repeatability across & Theory discovery, \\
%  &  & from observations & broad parameter spaces & boundary exploration \\
% \addlinespace
% Deductive & $(T, C) \rightarrow O$ & Test theoretical & Multi-variable control, & Policy simulation, \\
%  &  & predictions & simultaneous hypothesis competition & hypothesis testing \\
% \addlinespace
% Abductive & $(O, T) \rightarrow C$ & Infer mechanisms, & Systematic mechanism space search, & Mechanism discovery, \\
%  &  & then validate & precise hypothesis generation & behavioral explanation \\
% \bottomrule
% \end{tabular}
% \end{table}

\begin{figure}[t!]
    \centering
    \setlength{\fboxrule}{0.pt}
    \setlength{\fboxsep}{0.pt}
    \includegraphics[width=1\linewidth]{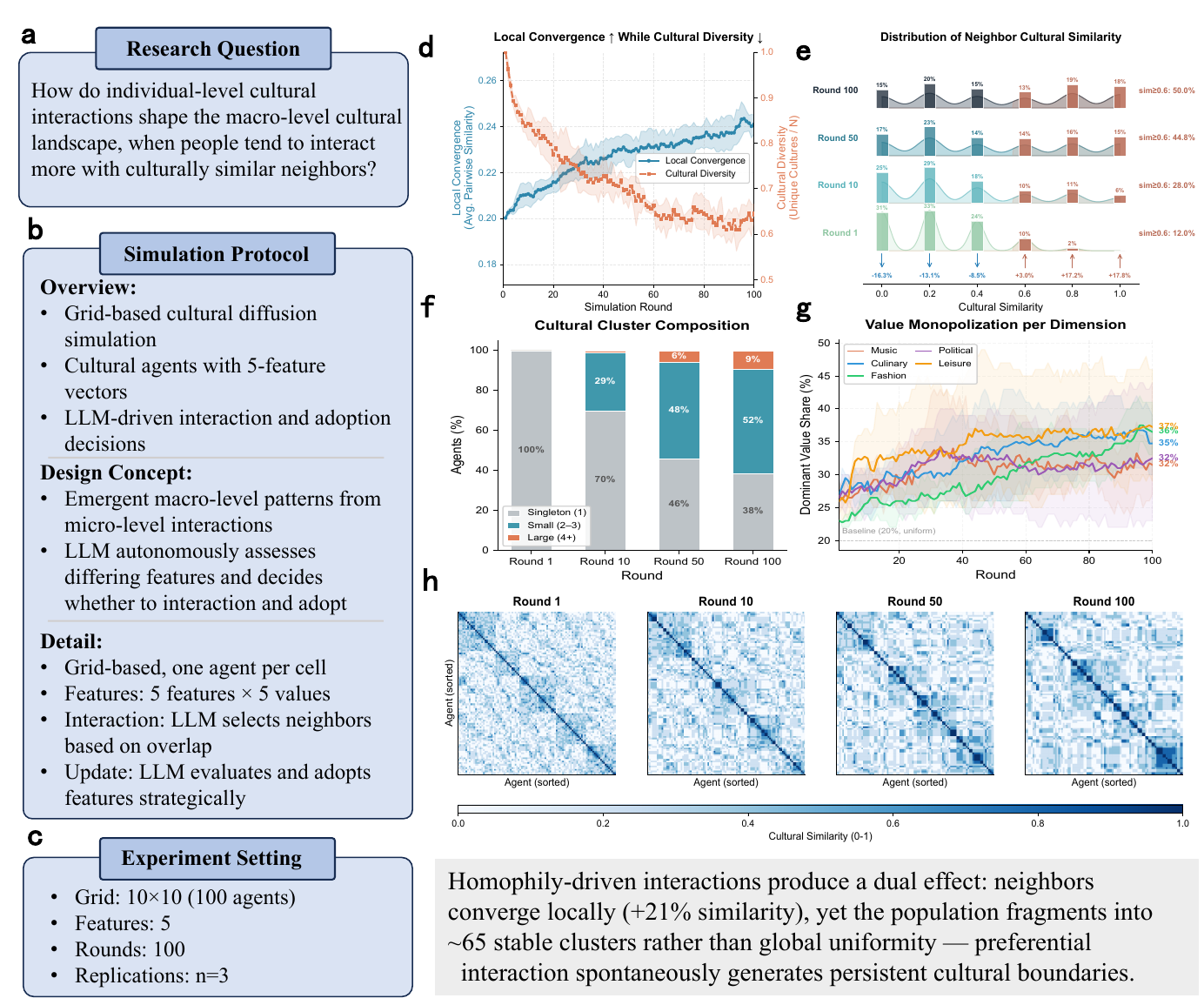}
    \caption{\textbf{Inductive paradigm: S-Researcher autonomously 
reproduces Axelrod's cultural dissemination dynamics, confirming 
coexistence of local convergence and global polarization.}
\textbf{a}, Research question input to S-Researcher.
\textbf{b}, Simulation protocol following the ODD standard.
\textbf{c}, Experimental setup: 100 LLM agents on a $10\times10$ 
grid, 5 cultural features $\times$ 5 values, 100 rounds, 3 replicates.
\textbf{d}, Dual-metric time series: local convergence (blue, left 
axis) increases from $\sim$0.20 to $\sim$0.24 (+21.0\%); cultural 
diversity (red, right axis) decreases from $\sim$1.0 to $\sim$0.65. 
Shaded areas show variability across replicates.
\textbf{e}, Distribution of neighbor cultural similarity across 
rounds, with cumulative high-similarity proportion (sim $\geq$ 0.6) 
increasing from 12.0\% to 50.0\%.
\textbf{f}, Cultural cluster composition over time: singleton agents 
decline while large clusters ($4+$) grow from 0\% to 38\%.
\textbf{g}, Dominant value share evolution across five cultural 
dimensions, rising from the uniform baseline (20\%) to a range of 32-37\%.
\textbf{h}, Pairwise cultural similarity matrices at four time points 
(Rounds 1, 10, 50, 100), showing progressive sharpening of cultural 
boundaries.
}
    \label{fig:inductive}
\end{figure}

\subsection*{Reproducing cultural dissemination dynamics (Inductive)}

Inductive research seeks to discover generalizable patterns from 
observations without presupposing a theory. A key challenge is that 
emergent macro-level phenomena often arise from micro-level interactions 
across large populations, requiring experiments at scales that are 
difficult to achieve with human participants alone. S-Researcher 
addresses this by enabling researchers to deploy large-scale LLM agent 
simulations under arbitrary scenario descriptions and automatically 
identify emergent regularities from agent interactions.

As a test case, we task S-Researcher with investigating how individual-level cultural interactions shape the macroscopic social landscape. The sole input is a natural-language research question — specifically, how cultural interactions among LLM-driven agents give rise to macro-level social dynamics (Fig.~\ref{fig:inductive}a) — accompanied by a brief scenario description. From this minimal input alone, S-Researcher autonomously designs a complete simulation protocol (Fig.~\ref{fig:inductive}c), deploying 100 agents on a 10×10 grid. Each agent is initialized with five cultural feature dimensions — music preference, dietary habit, fashion style, political orientation, and leisure activities — each taking one of five possible values (Fig.~\ref{fig:inductive}b). Agent interactions follow the homophily principle, whereby neighbors with greater cultural similarity are assigned higher interaction probabilities. S-Researcher then executes the simulation over 100 rounds across three independent replicates. Critically, no established social theories are provided to guide agent behavior at any stage of the process.

After collecting the results, S-Researcher discovers that: over 100 rounds, average pairwise cultural similarity increases by 21.0\%, from ${\sim}$0.20 to ${\sim}$0.24 (Fig.~\ref{fig:inductive}d, blue curve), indicating progressive local convergence. Yet global cultural uniformity does not emerge: cultural diversity declines from ${\sim}$1.0 to only ${\sim}$0.65, meaning the population stabilizes into approximately 65 distinct cultural ``islands'' (Fig.~\ref{fig:inductive}d, red curve). Pairwise similarity matrices illustrate this transition visually—cultural boundaries sharpen from a near-uniform distribution in Round 1 to clearly delineated block structures by Round 100 
(Fig.~\ref{fig:inductive}h). Ridge plots reveal the underlying microscale mechanism: the proportion of high-similarity neighbor pairs (sim $\geq$ 0.6) increases from 12.0\% to 50.0\%, producing a bimodal distribution of within-group homogeneity and between-group heterogeneity (Fig.~\ref{fig:inductive}e). All five cultural dimensions simultaneously undergo convergence, with dominant value shares rising from the uniform baseline of 20\% to 32–37\% (Fig.~\ref{fig:inductive}g).

The above observations—local assimilation coexisting with persistent global diversity—precisely recapitulate the central paradox of Axelrod's classical cultural dissemination model~\cite{axelrod1997dissemination}. 
S-Researcher's auto-generated report independently identifies this theory and the underlying mechanism: homophily drives culturally similar agents toward further convergence, but once inter-group cultural distance exceeds a critical threshold, interaction ceases entirely, forming impassable cultural boundaries. 
Crucially, these dynamics emerge from language model reasoning rather than hard-coded behavioral rules—validating S-Researcher's capacity to autonomously design experiments and rediscover established social phenomena through pure induction.

\begin{figure}[t!]
    \centering
    \setlength{\fboxrule}{0.pt}
    \setlength{\fboxsep}{0.pt}
    \fbox{
        \includegraphics[width=\linewidth]{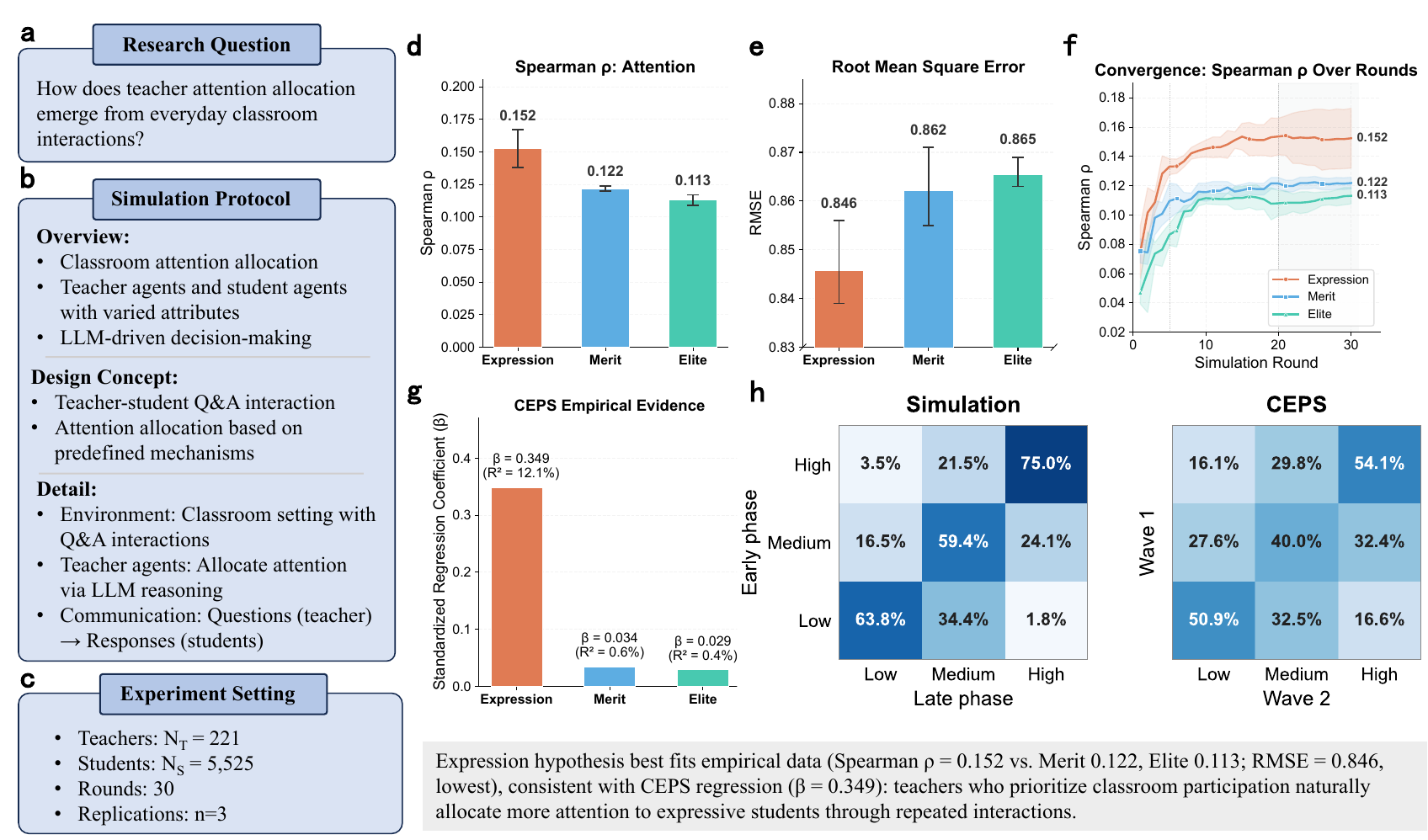}
    }
    \caption{\textbf{Deductive paradigm: bottom-up classroom simulation independently recovers the empirically established dominance of expressive ability in teacher attention allocation.}
\textbf{a}, Research question input to S-Researcher.
\textbf{b}, Simulation protocol following the ODD standard.
\textbf{c}, Experimental setup: 221 simulated classrooms, 5,525 student agents with profiles from CEPS, 30 rounds $\times$ 3 replicates.
\textbf{d}, Spearman $\rho$ between simulated and empirical CEPS attention rankings ($\uparrow$): Expression ($0.152$) $>$ Merit ($0.122$) $>$ Elite ($0.113$).
\textbf{e}, Root-mean-square error ($\downarrow$): Expression ($0.846$) yields the best fit.
\textbf{f}, Convergence of $\rho$ over 30 simulation rounds; hypothesis ranking stabilizes after round~5.
\textbf{g}, Independent validation via CEPS regression: communicative ability ($\beta = 0.349$, $R^2 = 12.1\%$) explains substantially more variance than academic achievement or SES, confirming the simulation-derived ordering.
\textbf{h}, Attention dynamics comparison: transition matrices from simulation (early vs.\ late phase) and CEPS two-wave panel data (Wave~1 vs.\ Wave~2) both exhibit diagonal-dominant persistence; simulation reproduces the qualitative structure but overestimates rigidity (e.g., Low$\to$High: $1.8\%$ vs.\ $16.6\%$), consistent with its role as a mechanism-isolating experiment.
}
    \label{fig:deductive}
\end{figure}

\subsection*{Competing hypotheses on teacher attention allocation (Deductive)}

Deductive research tests theoretical predictions by systematically comparing competing hypotheses against empirical evidence. In traditional practice, however, simultaneously testing multiple hypotheses under controlled conditions requires recruiting large participant pools and implementing parallel experimental designs—a process that is both resource-intensive and logistically challenging. S-Researcher addresses this by encoding competing hypotheses as distinct simulation configurations and executing them in parallel at scale, enabling systematic hypothesis competition that would be difficult to achieve with human participants alone.

As a test case, we task S-Researcher with a critical question in educational equity: what determines teacher attention allocation in classrooms? After receiving this research question, S-Researcher automatically generates three competing hypotheses: the \textit{Expression Hypothesis}, which posits that communicative ability drives attention acquisition; the \textit{Merit Hypothesis}, which attributes attention to academic achievement; and the \textit{Elite Hypothesis}, which argues that family socioeconomic status plays the dominant role (Fig.~\ref{fig:deductive}a).
Three parallel simulations are then launched, one per hypothesis. Each simulation comprises 221 classrooms, each containing one teacher agent and approximately 25 student agents, 5,525 students total, with all student profiles drawn from CEPS.
After collecting the simulation results, S-Researcher computes the correlation between simulated and observed attention patterns in CEPS to identify the most plausible hypothesis. 
In attention ranking consistency, the Expression Hypothesis achieves a Spearman correlation of $\rho = 0.152$, significantly exceeding the Merit Hypothesis ($\rho = 0.122$) and Elite Hypothesis ($\rho = 0.113$; Fig.~\ref{fig:deductive}d).
From a dynamic perspective, the relative ranking of the three hypotheses stabilized after round 5 and remained consistent throughout the subsequent 25 rounds (Fig.~\ref{fig:deductive}f), demonstrating high reproducibility.
In prediction accuracy, the Expression Hypothesis yields the lowest root-mean-square error (RMSE = 0.846 vs.\ 0.862 and 0.865; Fig.~\ref{fig:deductive}e).
To provide independent supporting evidence, we examine whether real-world associations in the CEPS data are directionally consistent with S-Researcher's simulation results.
As shown in Fig.~\ref{fig:deductive}g, CEPS regression analysis confirms this consistency: the standardized coefficient for communicative ability ($\beta = 0.349$, $R^2 = 12.1\%$) explains approximately 20 times more variance in teacher attention than academic achievement ($\beta = 0.034$, $R^2 = 0.6\%$) or socioeconomic status ($\beta = 0.029$, $R^2 = 0.4\%$), corroborating the simulation's identification of the Expression Hypothesis as the dominant factor.
Although regression and simulation converge on the same conclusion, they differ in explanatory depth~\cite{epstein1999agent}: while regression identifies the most significant variables, simulation illustrates the underlying process whereby expressive students progressively accumulate attention advantages through round-by-round interactions.
This process-level interpretability, generated automatically, is a distinctive strength of simulation-based research that statistical association alone cannot provide.

\subsection*{Mechanism discovery and human validation in public goods games (Abductive)}

\begin{figure}[t!]
    \centering
    \setlength{\fboxrule}{0.pt}
    \setlength{\fboxsep}{0.pt}
    \fbox{
        \includegraphics[width=\linewidth]{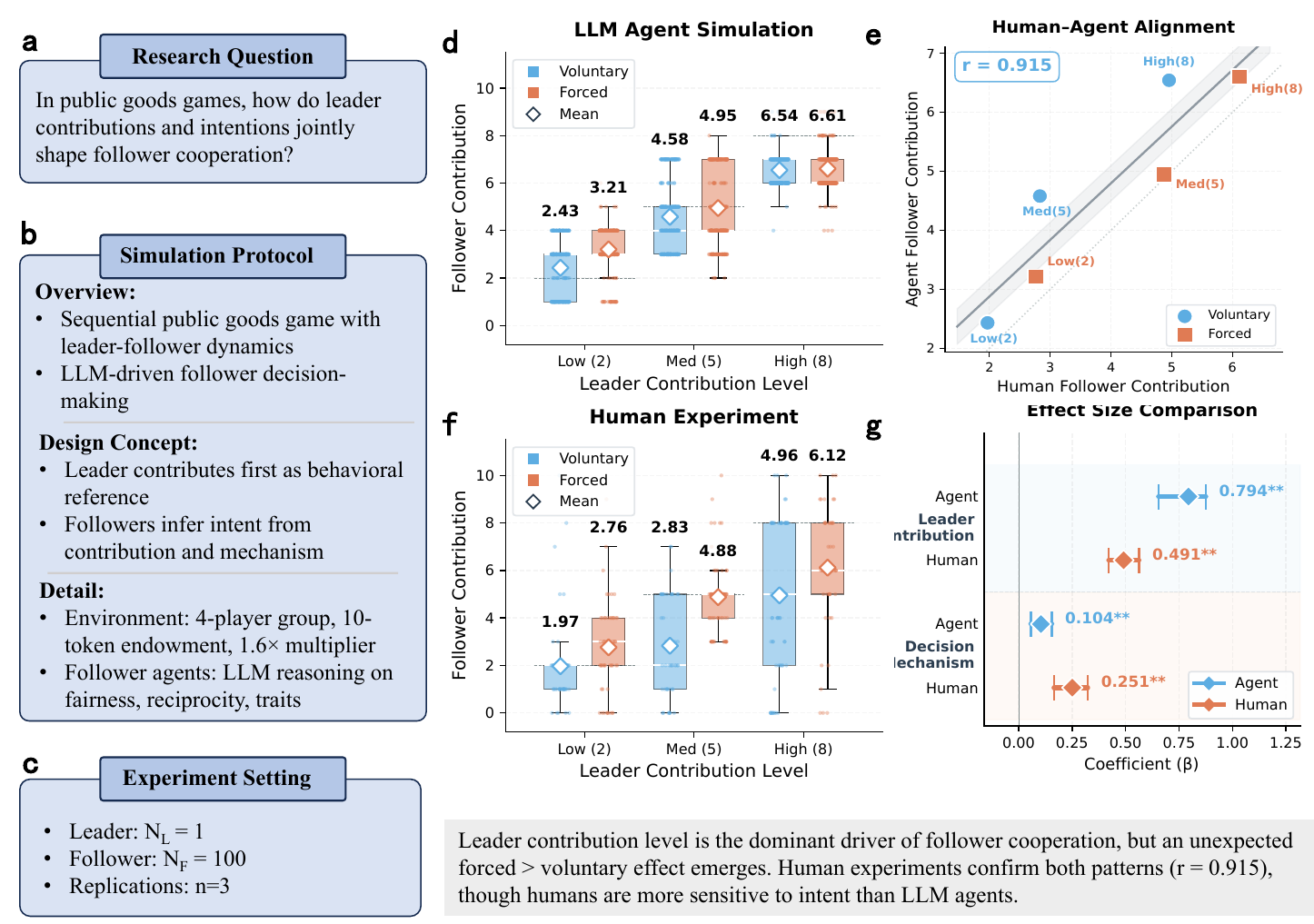}
    }
\caption{\textbf{Abductive paradigm: counterfactual decomposition of follower cooperation in public goods games reveals behavioral anchoring as the dominant mechanism and uncovers an unexpected forced~$>$~voluntary effect.}
\textbf{a}, Research question: what causal mechanisms drive follower cooperation when leaders contribute first?
\textbf{b}, Simulation protocol following the ODD standard.
\textbf{c}, Experimental setup: $2 \times 3$ between-subjects design (voluntary/forced $\times$ low/medium/high contribution), 100 agent followers per condition, 3 replicates; parallel human experiment ($N = 120$).
\textbf{d}, LLM agent simulation box plots ($N_F = 100$, $n = 3$): follower contributions under voluntary (blue) and forced (orange) conditions across leader contribution levels; diamonds mark means. Forced conditions elicit higher contributions at every level.
\textbf{e}, Human-agent alignment scatter plot across 6 conditions, Pearson $r = 0.915$.
\textbf{f}, Human experiment box plots ($N = 120$): humans show the same forced~$>$~voluntary pattern, with the gap widening at medium and high contribution levels.
\textbf{g}, Effect size comparison forest plot: both agents and humans show significant effects of leader contribution ($\beta_{\text{agent}} = 0.794$, $\beta_{\text{human}} = 0.491$) and decision mechanism ($\beta_{\text{agent}} = 0.104$, $\beta_{\text{human}} = 0.251$), with leader contribution consistently the dominant factor.}
    \label{fig:abductive}
\end{figure}

Abductive reasoning works backward from an observed phenomenon to uncover its underlying causal mechanisms. Unlike deductive approaches that test pre-specified hypotheses, abductive research asks: \textit{why does this phenomenon occur, and through what mechanisms?} This is naturally addressed through counterfactual experimentation, where systematically isolating and altering specific variables decomposes a complex phenomenon into its core drivers. S-Researcher operationalizes this paradigm by translating open-ended questions into targeted experimental manipulations, executing the simulations, and synthesizing the results to reveal the fundamental mechanisms at play.

Public goods games are a canonical paradigm in experimental economics where group members independently decide how much of their private endowment to contribute to a shared pool that benefits everyone~\cite{fehr2002altruistic,nowak2006five,szolnoki2010reward}. We task S-Researcher with a specific question in this setting: in leader--follower interactions~\cite{eichenseer2023leading}, how does the interplay between the leader's contribution amount and the perceived intent behind that contribution shape followers' cooperation decisions? (Fig.~\ref{fig:abductive}a).
After receiving this research question, S-Researcher automatically designs a $2 \times 3$ between-subjects experiment (Fig.~\ref{fig:abductive}b): leader decision mechanism (voluntary choice vs.\ forced assignment) $\times$ contribution level (low: 2 tokens, medium: 5 tokens, high: 8 tokens), yielding 6 treatment conditions. The game uses a classical four-player public goods framework (10-token endowment, $1.6\times$ multiplier), where follower agents observe both the leader's contribution amount and decision mechanism before making their cooperation decisions.
S-Researcher then executes the simulation with 100 follower agents per condition, replicated 3 times.

The simulation results generated by S-Researcher provide a clear mechanistic answer to how these factors interplay (Fig.~\ref{fig:abductive}d). The analysis reveals that follower agents' cooperation is predominantly anchored by the leader's absolute contribution level (the \textit{what}). However, the leader's underlying intent (the \textit{why}) serves as a statistically significant secondary modulator ($\beta = 0.104^{**}$, Fig.~\ref{fig:abductive}g). Notably, at equivalent contribution levels, follower agents contribute more when the leader's action is forced rather than voluntary. This dynamic suggests a potential intent-based penalty: agents may negatively attribute a \textit{voluntary} sub-optimal contribution to selfishness, thereby reducing their cooperation, whereas a \textit{forced} assignment removes personal culpability and yields a relatively higher baseline of reciprocity.

To validate these findings, we conduct parallel human experiments ($N = 120$, 3 rounds) using the identical setup (Fig.~\ref{fig:abductive}f). The empirical data corroborates the mechanistic structure predicted by the simulation, achieving a robust macroscopic alignment ($r = 0.915$, Fig.~\ref{fig:abductive}e). Crucially, human participants exhibit the exact same directional trends: they prioritize the observable action ($\beta = 0.491^{**}$) over the decision mechanism ($\beta = 0.251^{**}$), and similarly display higher cooperation in forced scenarios. While the simulated trajectory successfully mirrors this empirical reality, decomposing the effect sizes (Fig.~\ref{fig:abductive}g) exposes a divergence in the magnitude of cognitive weightings. Although sharing the same decision-making framework, LLM agents heavily over-index on the absolute contribution amount, whereas human followers demonstrate a significantly higher sensitivity to the underlying intent than LLM.

\subsection*{Expert evaluation across three paradigms}

\begin{figure}[t!]
    \centering
    \setlength{\fboxrule}{0.pt}
    \setlength{\fboxsep}{0.pt}
    \fbox{
        \includegraphics[width=\linewidth]{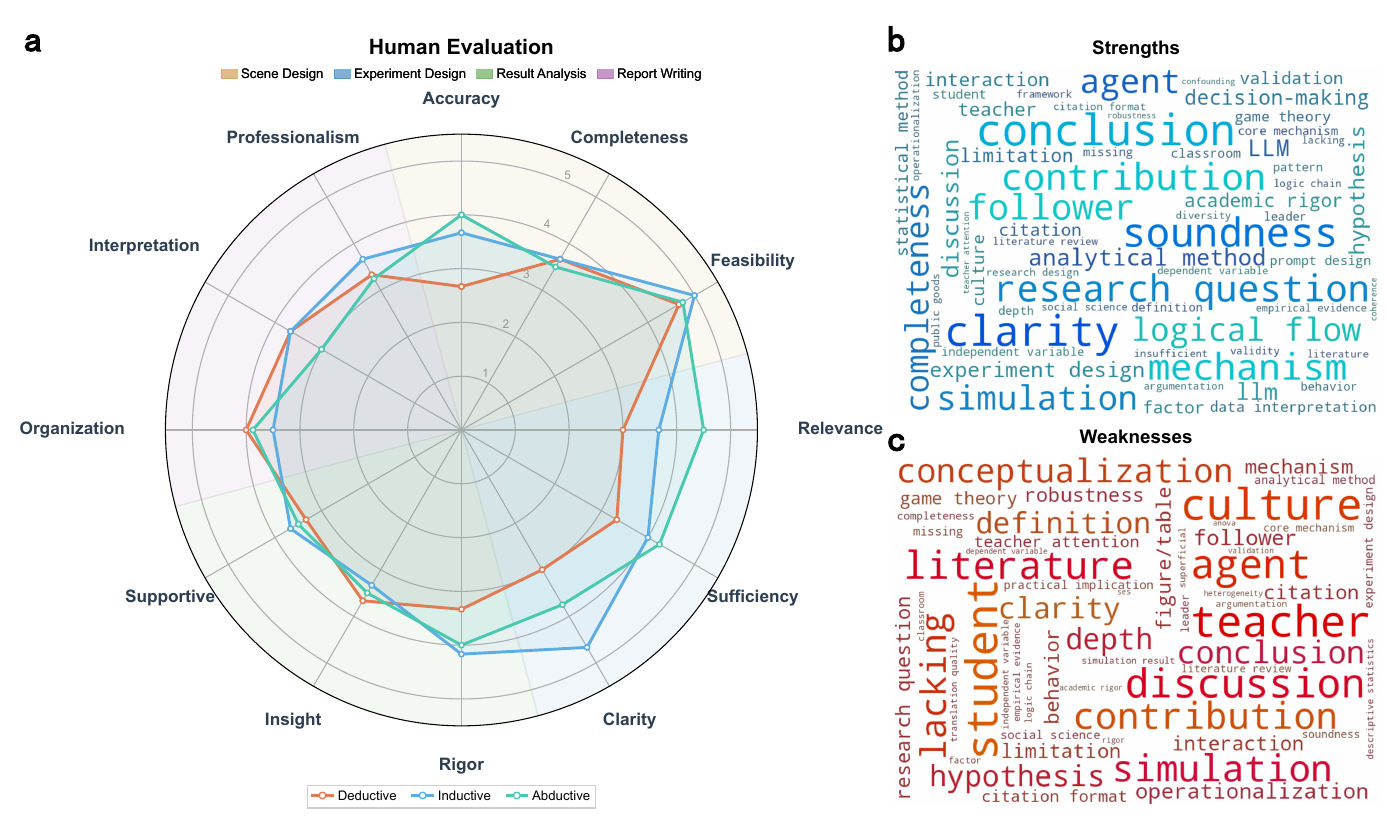}
    }
    \caption{\textbf{Cross-paradigm expert evaluation of S-Researcher outputs.}
\textbf{a}, Radar chart comparing expert scores across 12 evaluation dimensions (grouped into four categories: Scene Design, Experiment Design, Result Analysis, and Report Writing) for each experiment paradigm. Scores are on a 1--5 Likert scale (3 reviewers per paradigm). The inductive paradigm (3.89/5) and abductive paradigm (3.82/5) receive higher overall ratings than the deductive paradigm (3.47/5). All three paradigms score highest on Feasibility and Relevance, while Accuracy and Insight show the most room for improvement.
\textbf{b}, Word cloud of reviewer-identified strengths: dominant themes include sound conclusions, clear research questions, logical flow, and appropriate analytical methods.
\textbf{c}, Word cloud of reviewer-identified weaknesses: recurring concerns include insufficient literature engagement, shallow discussion, lack of conceptual depth, and incomplete operationalization of key constructs.}
    \label{fig:evaluation}
\end{figure}

To assess the scientific quality of S-Researcher's automated outputs, we recruited three independent domain experts per paradigm, all with active research backgrounds in the social sciences and humanities, to evaluate the complete auto-generated research reports. Each expert reviewed the full report independently, without access to other reviewers' scores. Reviewers scored 12 dimensions on a 1--5 Likert scale, organized into four categories: Scene Design (accuracy, completeness, feasibility), Experiment Design (relevance, sufficiency, clarity), Result Analysis (rigor, insight, supportiveness), and Report Writing (organization, interpretation, professionalism).

Figure~\ref{fig:evaluation}a reveals a consistent cross-paradigm  pattern. The inductive paradigm receives the highest overall rating (3.89/5), followed by the abductive (3.82/5) and deductive (3.47/5) paradigms. All three paradigms score well on Feasibility and Relevance ($>$3.5), indicating that S-Researcher reliably generates technically implementable designs aligned with the research questions. Accuracy and Insight show the most room for improvement across all paradigms. The lower Accuracy scores reflect the inherent difficulty of translating abstract scenario descriptions into concrete simulation configurations that faithfully capture real-world social mechanisms across a vast design space which is a challenge that demands deep domain knowledge even for human researchers. Similarly, the lower Insight scores highlight the difficulty of moving beyond descriptive summaries to generate theoretically grounded interpretations that connect simulation results to broader social science knowledge, a higher-order analytical capability that remains demanding for both LLMs and junior human researchers alike. The deductive paradigm's lower overall score is primarily attributable to weaker performance in Result Analysis and Report Writing: interpreting multi-hypothesis comparisons with modest effect sizes demands deeper contextual reasoning than the clearer signal-to-noise ratios in the inductive (emergent macro-patterns) and abductive (near-linear dose--response) studies, exposing a current limitation of LLM-based analytical reasoning under ambiguity.

Qualitative analysis of reviewer comments (Fig.~\ref{fig:evaluation}b,c) identifies systematic strengths and weaknesses. Reviewers consistently praised the logical flow and soundness of conclusions, the clarity of research questions, and the appropriate use of analytical methods, all of which reflect the structured, multi-agent pipeline underlying S-Researcher's automation. The most frequent criticisms concerned insufficient engagement with existing literature, shallow theoretical discussion, and the absence of robustness checks. These weaknesses are not unique to S-Researcher but reflect broader limitations of current LLM-generated academic writing, where deep domain grounding and critical self-reflection remain challenging. They point to concrete directions for improvement: integrating automated literature retrieval and citation into the report generation pipeline, incorporating domain-specific theoretical frameworks to deepen interpretive depth, and adding systematic sensitivity analyses across different backbone LLMs to establish result robustness.

\section*{Discussion}

\textbf{Capabilities and boundaries of LLM simulation.}
Our three case studies allow us to triangulate the current capabilities and limitations of LLM-based social simulation.
At the macroscopic level, LLM agents exhibit high credibility: in the inductive paradigm, the simulation faithfully reproduces classical cultural dynamics with emergent regularities arising from agent interactions rather than rule-based programming.
In multi-hypothesis competition (deductive paradigm), simulations produce directionally consistent conclusions that align with empirical survey data.
In mechanism discovery (abductive paradigm), agents generate testable hypotheses with primary effect sizes that quantitatively match human experimental data ($r = 0.915$).

However, systematic limitations are equally clear.
LLM agents exhibit substantially lower behavioral heterogeneity than humans: their response variability is 20--300 times smaller, and they produce more extreme distributional patterns.
They also show reduced sensitivity to social reasoning cues such as intentionality, as evidenced by the near-zero decision mechanism effect in the public goods game.
These characteristics suggest that while LLM agents are effective proxies for studying central tendencies and primary effects, research involving behavioral heterogeneity, intention sensitivity, or distributional tails should incorporate human participants.

\textbf{Organizing simulation around reasoning paradigms.}
We organized S-Researcher's automation around the three canonical modes of scientific reasoning~\cite{peirce1931collected}.
This organization is not a philosophical claim but a practical design choice: it provides a structured way to map diverse social science research questions onto appropriate simulation workflows.
The three case studies demonstrate that this organization covers a broad range of research types, from phenomenon reproduction and pattern identification (induction) through confirmatory hypothesis testing (deduction) to mechanism inference with human validation (abduction).
% In the method section, we provide a unified The $O = F(T,C)$ formalization (see Methods) provides a unified language for identifying which paradigm best fits a given research question.

\textbf{Limitations and future directions.}
Several limitations warrant acknowledgement.
First, while the human experiment ($N = 120$) independently validates the abductive paradigm's core finding, larger sample sizes would strengthen the generalizability of cross-paradigm human--agent alignment assessments.
Second, the three case studies have been primarily demonstrated on cognitive-reasoning-oriented behaviors; emotion-driven and culturally embedded behaviors may require additional considerations.
Third, expert reviews identified the absence of formal literature integration and variable operationalization in auto-generated reports, capabilities that future versions of S-Researcher should develop.
Looking forward, we envision expanding the scenario library to cover additional social science domains, further systematizing the human--AI collaborative research workflow, and developing methods to calibrate agent behavioral heterogeneity against known human population parameters.

\section*{Methods}\label{method}

\subsection*{YuLan-OneSim system architecture}

S-Researcher requires a simulation platform that is simultaneously general (capable of executing arbitrary experimental designs), scalable (supporting population-level experiments), and reliable (producing trustworthy behavioral outputs).
We developed YuLan-OneSim to meet these requirements through three integrated capabilities: an auto-programming framework for generality, a distributed simulation architecture for scalability, and a feedback mechanism for reliability.

\textbf{Auto-programming framework for generalizable simulation.}
To support arbitrary experimental designs, YuLan-OneSim translates natural-language scenario descriptions into executable simulation code through a structured pipeline.
User inputs are first formalized following the ODD (Overview, Design Concepts, Details) protocol~\cite{grimm2010odd}, a standard documentation format for agent-based models, through interaction with a dedicated LLM agent that resolves ambiguities.
A behavior graph $\mathcal{G} = \{\mathcal{N}, \mathcal{E}\}$ is then constructed, where each node $n_i \in \mathcal{N}$ represents an action associated with an agent type, and each directed edge $e_{ij} \in \mathcal{E}$ indicates that action $n_i$ can trigger action $n_j$ via a structured event carrying typed variables.
The graph undergoes structural validation and semantic validation.
Guided by the validated graph, executable code is generated via breadth-first traversal, instantiating each node against base code templates, and iteratively refined through compilation testing and logical verification.

Finally, code and data are decoupled: environment data, agent profiles, and agent relationships are specified independently to enable flexible configuration. This progressive construction from logical specification to structured behavior graph and finally to concrete code constitutes a workflow that ensures stable, high-quality scenario generation.
Ablation analysis confirms that the behavior graph is the most influential component: its removal leads to the largest quality degradation, followed by removing the code-level refinement module and the graph validation module.
To further support researchers, we provide a default repository of 50 scenarios spanning eight social-science domains, including economics, sociology, politics, psychology, organizational studies, demography, law, and communication, as ready-to-use examples.

\textbf{Distributed architecture for scalable simulation.}
Each agent in YuLan-OneSim comprises four modules: a \textit{profile} module storing public and private attributes, a \textit{memory} module supporting configurable storage strategies (e.g., sliding windows, vector databases, knowledge graphs), a \textit{planning} module implementing reasoning approaches such as Chain-of-Thought, Belief-Desire-Intention, and Theory-of-Mind, and an \textit{action} module that converts the above components into specific behaviors.
The simulation engine employs an event-driven, asynchronous architecture centered on an event bus: events encapsulate agent actions and environmental state changes with full contextual metadata, while agents subscribe to relevant event types, enabling natural representation of concurrent activities and complex causal relationships.

For large-scale simulations, a Master--Worker distributed design orchestrates parallel execution.
The master node manages the global agent registry and simulation state; worker nodes execute agent logic in parallel, communicating via gRPC with optimized message batching; topology-aware allocation co-locates frequently interacting agents on the same worker to minimize cross-node communication; and a proxy interface provides a consistent API for both distributed and non-distributed deployments, enabling seamless scaling.

\textbf{VR$^2$T feedback mechanism for reliable simulation.}
General-purpose LLMs are not specifically trained for social-science domains and often produce unreliable simulation outputs that can accumulate over extended runs.
To address this, we design a Verifier--Reasoner--Refiner--Tuner (VR$^2$T) multi-agent feedback framework that iteratively evaluates simulation outputs and fine-tunes the backbone LLMs.

For each prompt--response pair $(p, r)$ generated during simulation, the \textit{verifier} assesses response quality and assigns a score $s = V(p, r)$; the \textit{reasoner} generates an interpretable explanation $\xi = R(p, r, s)$; if $s$ falls below a predefined threshold $\theta$, the \textit{refiner} produces a corrected response $r' = F(p, r, s, \xi)$.
This process yields annotated tuples $\{(p_i, r_i, s_i, r'_i)\}_{i=1}^{N}$ for all underperforming samples.
The \textit{tuner} then fine-tunes the backbone LLM $\pi_\phi$ using one of two objectives.
For supervised fine-tuning (SFT), the model minimizes the negative log-likelihood over refined responses:
\begin{equation}
    \mathcal{L}_{\text{SFT}}(\phi) = -\mathbb{E}_{(p, r')} \left[ \log \pi_\phi(r' \mid p) \right].
\end{equation}
For direct preference optimization (DPO), the original and refined responses form a dispreferred--preferred pair, optimized via:
\begin{equation}
    \mathcal{L}_{\text{DPO}}(\phi) = -\mathbb{E}_{(p, r, r')} \left[ \log \sigma \!\left( \beta \log \frac{\pi_\phi(r' \mid p)}{\pi_{\text{ref}}(r' \mid p)} - \beta \log \frac{\pi_\phi(r \mid p)}{\pi_{\text{ref}}(r \mid p)} \right) \right],
\end{equation}
where $\pi_{\text{ref}}$ is the reference policy (pre-fine-tuning model), $\sigma(\cdot)$ is the sigmoid function, and $\beta$ controls the KL divergence constraint.
Each component can operate in either fully automatic or human-in-the-loop mode, enabling the platform to serve as a ``social-science gym'' in which LLMs progressively acquire domain-specific social intelligence through continual simulation and feedback.

\subsection*{S-Researcher architecture}

\textbf{Research paradigm formalization.}
Following Peirce's classical trichotomy of scientific reasoning, we organize S-Researcher around three complementary research paradigms that together span the full methodological spectrum of social science inquiry.
\begin{itemize}

\item \textit{Inductive paradigm}: this paradigm operates without presupposing a theoretical framework. The system executes large-scale simulations across broad parameter spaces and applies statistical pattern discovery to identify regularities emerging from agent interactions. The unique simulation advantage lies in achieving scale and repeatability that are difficult to attain with human participants, enabling researchers to explore vast behavioral landscapes and uncover macro-level patterns from micro-level dynamics. Typical applications include theory discovery and boundary exploration~\cite{heit2000properties,helbing2012agent,bonabeau2002agent}.

\item \textit{Deductive paradigm}: rooted in the hypothetico-deductive method~\cite{popper1959logic}, this paradigm begins with competing theoretical hypotheses, derives testable predictions under controlled conditions, and evaluates which predictions best match empirical observations. S-Researcher encodes each hypothesis as a distinct simulation configuration and executes them in parallel, enabling systematic multi-hypothesis competition that would be prohibitively resource-intensive with human participants alone. Typical applications include policy simulation and confirmatory hypothesis testing.

\item \textit{Abductive paradigm}: this paradigm works backward from an observed but unexplained phenomenon to uncover its underlying causal mechanisms. The system generates candidate mechanisms and designs targeted counterfactual experiments to discriminate among them, systematically isolating and manipulating specific variables to decompose complex phenomena into core drivers. Typical applications include mechanism discovery and behavioral explanation.
\end{itemize}
These three paradigms provide a unified organizational language for mapping diverse social science research questions onto appropriate simulation workflows: when competing theories exist, deduction tests them head-to-head; when no prior theory is available, induction discovers regularities from observations; when a phenomenon demands causal explanation, abduction generates and validates candidate mechanisms. A \textit{paradigm selection agent} automatically analyzes the user-provided research context, comprising the research question, available theories, and observational evidence. It then recommends the most appropriate paradigm by evaluating diagnostic signals; for instance, the presence of competing theories favors deduction, while unexplained observations favor abduction. Users may override this recommendation at any point.

\textbf{Experiment design module.}
The experiment design pipeline comprises three sequentially orchestrated LLM agents that transform a user-provided research question and scenario description into a complete, executable experimental design.
First, a \textit{detailer agent} elaborates the user's scenario into a complete ODD protocol, specifying agent types, interaction patterns, communication protocols, decision mechanisms, and domain-specific behavioral constraints in a structured format that can be directly consumed by YuLan-OneSim's auto-programming framework.
Second, an \textit{experiment configuration agent} generates the experimental design, defining control and treatment groups, replication settings (number of replicates, seed strategy), and analysis configuration (statistical tests, confidence levels, comparison groups). This agent separates the experimental logic from the simulation implementation, ensuring that the statistical design is well-powered and methodologically sound regardless of the underlying simulation complexity.
Third, an \textit{intervention agent} produces fine-grained profile modifications for each treatment group. For deductive research, each treatment group implements a distinct hypothesis through unique agent decision logic; for abductive research, the agent generates systematic condition manipulations across multiple levels of the independent variable.
Target agents for intervention are selected via configurable strategies (by agent type or by profile criteria with percentage thresholds), and modifications are applied either as direct value assignments or through LLM-based profile enhancement that uses natural-language templates to ensure behavioral coherence after modification.

\textbf{Results analysis module.}
The raw outputs of the simulation system, which comprise per-agent and per-round records across multiple replicates and treatment conditions, are typically too large and complex to interpret directly.
We implement a multi-agent analytical framework in which specialized agents collaborate to progressively distill raw data into scientific insights through three phases.
In the \textit{data collection} phase, a data processing agent aggregates simulation runs across experimental groups, extracts per-step metrics, and generates an LLM-based semantic summary that characterizes the overall data landscape.
In the \textit{planning} phase, an analysis planner agent examines this summary alongside metric profiles, automatically classifying variables by data type and granularity to produce a structured analysis plan. This plan specifies the research questions, analysis types, and target metrics for each investigation.
In the \textit{analysis} phase, a team of agents collaboratively executes the plan: a visualization agent translates each plan item into figure specifications, generates and executes plotting code in a sandboxed environment; a statistical analysis agent selects appropriate methods from a registry of over 40 tools (spanning parametric, non-parametric, time-series, Bayesian, and multi-factor designs) based on data characteristics and the research question; and a review agent evaluates the resulting figures and statistical outputs, providing iterative feedback to refine the analyses until the findings are coherent and well-supported.

\textbf{Report generation module.}
The report generation module converts accumulated research artifacts, spanning experimental design, simulation data, and statistical analyses, into a structured academic report through multi-agent collaboration.
An \textit{outline agent} structures the report following standard academic conventions, with paradigm-aware content requirements: deductive reports emphasize comparative hypothesis validation, inductive reports highlight emergent patterns, and abductive reports foreground causal mechanisms and dose--response relationships.
A \textit{writing agent} then produces each section independently using paradigm-specific prompts, automatically incorporating figures and statistical results from the analysis module.
A \textit{review agent} evaluates the assembled report across four quality dimensions, namely technical rigor, clarity, validation, and writing quality, with each category scored on a 0--5 scale.
Sections scoring below quality thresholds are automatically revised and reassembled in an iterative generation--review cycle until all criteria are satisfied.

\bibliographystyle{unsrt}
\bibliography{references}

% Extended Data Figures or Tables
\clearpage
\setcounter{figure}{0}
\renewcommand{\figurename}{Extended Data Figure}
\setcounter{table}{0}
\renewcommand{\tablename}{Extended Data Table}

% add extended data figures/tables below (up to 6 items)

\end{document}